\documentclass{bmvc2k}
\usepackage{times}
\usepackage{graphicx}
\usepackage{amsmath}
\usepackage{amssymb}
\usepackage{adjustbox}
\usepackage{multirow}
\usepackage{multicol}
\usepackage{booktabs}
\title{Teacher-Class Network: A Neural Network Compression Mechanism}

\addauthor{Shaiq Munir Malik}{18030012@lums.edu.pk}{1}
\addauthor{Mohbat Tharani}{mohbaf@rpi.edu}{2}
\addauthor{Muhammad Umair Haider}{m.haider@lums.edu.pk}{1}
\addauthor{Muhammad Musab Rasheed}{19030008@lums.edu.pk}{1}
\addauthor{Murtaza Taj}{murtaza.taj@lums.edu.pk}{1}

\addinstitution{
 Computer Vision and Graphics Lab\\
 Lahore Univ. of Management Sciences\\
 Lahore, Pakistan\\
 https://cvlab.lums.edu.pk/
}
\addinstitution{
 Rensselaer Polytechnic Institute\\
 New York, USA
}

\runninghead{Shaiq M. Malik, Murtaza Taj}{}

\begin{document}

\maketitle

%%%%%%%%% ABSTRACT
\begin{abstract}
To reduce the overwhelming size of Deep Neural Networks, \emph{teacher-student} techniques aim to transfer knowledge from a complex teacher network to a simple student network. We instead propose a novel method called the \emph{teacher-class} network consisting of a single teacher and multiple student networks (class of students). Instead of transferring knowledge to one student only, the proposed method divides learned space into sub-spaces, and each sub-space is learned by a student. Our students are not trained for problem-specific logits; they are trained to mimic knowledge (dense representation) learned by the teacher network; thus, the combined knowledge learned by the \emph{class of students} can be used to solve other problems. The proposed \emph{teacher-class} architecture is evaluated on several benchmark datasets such as MNIST, Fashion MNIST, IMDB Movie Reviews, CIFAR-10, and ImageNet on multiple tasks such as image and sentiment classification. Our approach outperforms the state-of-the-art single student approach in terms of accuracy and computational cost while achieving a $10-30$ times reduction in parameters. Code is available at \url{https://github.com/musab-r/TCN}.

%\keywords{Model Compression, Teacher-Student Network, Convolution Neural Networks}

% model compression; knowledge distillation; teacher-student network; 
\end{abstract}
\vspace{-.6cm}

\begin{figure}[t]
\centering
\begin{center}
   %\includegraphics[width=0.8\linewidth]{egfigure.eps}
\end{center}
\includegraphics[width=0.9\textwidth]{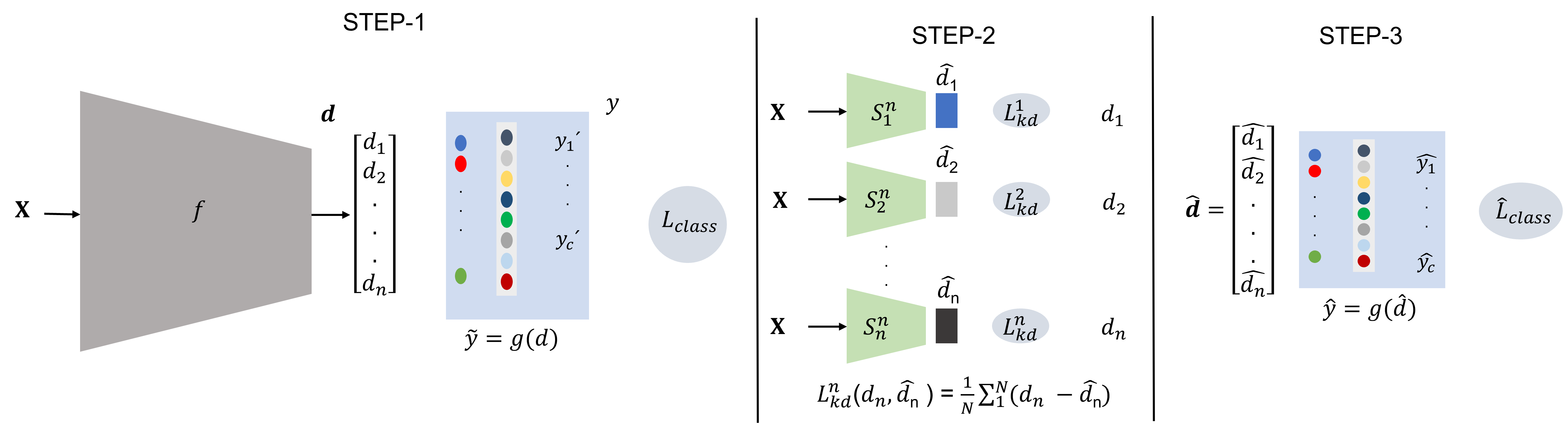}
   %\vspace*{3mm}
   \caption{Process overview: The dense feature $\mathbf{d}$ learned by the teacher network is divided into sub-spaces $d_i$. Each sub-space is learned by an individual student. Finally, the knowledge from all students is merged and fed to an output layer for the final decision.}
    \label{fig:overview}
    \vspace{-0.3cm}
\end{figure}

%%%%%%%%% BODY TEXT
\section{Introduction}
%\vspace{-0.2cm}
% ~\cite{chen2003application,dahl2011context,ferreira2018classification,taylor2002neural,ubeyli2009adaptive,ubeyli2010automatic}
Deep neural networks have effectively tended to various real-world problems, e.g., image classification~\cite{he2019bag,algan2021image}, visual detection and segmentation~\cite{minaee2021image,zhang2021weakly}, and audio recognition and analysis~\cite{purwins2019deep,morrone2021audio}. The availability of a large amount of training data, compute power, and improved deep neural network architectures~\cite{he2016deep,krizhevsky2012imagenet,simonyan2014very,szegedy2016rethinking} have enabled the deep learning domain to enhance its accuracy continuously. However, these networks have massive parameters and are resource-heavy; hence deploying such networks on resource deficient devices such as mobile phones is almost impractical. Subsequently, compact models with comparable accuracy are critically required.

There have been several efforts to compress these networks, such as efficient architectural blocks (separable convolution~\cite{Yoo_2018} and pyramid pooling~\cite{Xiang_2019}), pruning layers and filters~\cite{chen2018shallowing,he2019filter,alqahtani2021pruning,mohbat2019dimensionality, HaiderICIP2021}, quantization~\cite{Nogami_2019,Wu_2016}, and knowledge distillation~\cite{hinton2015distilling,lee2018self,passalis2018learning,watanabe2017student,guo2020online,you2018learning}. Efficient architectural blocks and pruning schemes make the model smaller without any reduction in the complexity of the problem, thus resulting in degraded performance~\cite{lee2018self}. Similarly, quantization causes loss of data due to approximation resulting in performance drop~\cite{Wu_2016}. 

Teacher(s)-student architecture~\cite{hinton2015distilling} uses a huge pre-trained network (teacher) to train a small model (student), so it can learn the knowledge extracted by the teacher that the student otherwise would not be able to learn because of its simpler architecture and fewer number of parameters. This knowledge transfer is achieved by minimizing the loss between the soft labels (probabilities produced by the softmax at a higher temperature~\cite{hinton2015distilling}) produced by the teacher and the student.

This paper proposes a novel neural network compression methodology called \emph{teacher-class} networks. As compared to existing literature~\cite{du2020agree,hinton2015distilling,lee2018self,mirzadeh2019improved,passalis2018learning,wang2018adversarial,watanabe2017student}, our proposed architecture has two key differences, (i)~instead of just one student, the proposed architecture employs multiple students to learn mutually exclusive chunks of the knowledge and (ii) instead of training student on the soft labels (probabilities produced by the softmax) of the teacher, our architecture tries to learn dense feature representations, thus making the solution problem independent. The size of chunks (sub-space) each student has to learn depends on the number of students. {{Unlike the ensemble-based/multi-branch method~\cite{NIPS2018_7980}, all of the students in our proposed approach have been trained independently, thus allowing model parallelism while reducing compute and memory requirements. The knowledge learned by each individual student in our case is combined and output layers are applied.}} These layers can be borrowed from the teacher network with pre-trained weights and can also be fine-tuned to further improve the loss occurred while transferring the knowledge to students. An overview of the proposed methodology is demonstrated in Fig.~\ref{fig:overview}.

\section{Related Work}
\textbf{Knowledge distillation via single student}: 
In knowledge distillation, as introduced by Hinton \emph{et al.}~\cite{hinton2015distilling}, a single student either tries to mimic a single teacher's~\cite{zhu2021complementary,liu2020adaptive,li2020hierarchical,du2020agree,hinton2015distilling,lee2018self,mirzadeh2019improved,passalis2018learning,wang2018adversarial,watanabe2017student, NIPS2018_7980} or multiple teachers~\cite{You_2019, guo2020online, Yang_2020}. Most of such schemes transfer knowledge to student by minimizing the error between the knowledge of the teacher and the student~\cite{hinton2015distilling,lee2018self}. Rather than matching actual representation, Passalis \emph{et al.}~\cite{passalis2018learning} and Watanabe \emph{et al.}~\cite{watanabe2017student} propose to model the knowledge using probability distribution and then match the distribution of teacher and student networks. Nikolaos \emph{et al.}~\cite{passalis2018learning} try to cater non-classification problems in addition to classification problems. Wang \emph{et al.}~\cite{wang2018adversarial} argue that it is hard to figure out which student architecture is more suitable to quantify the information inherited from teacher networks, so they use generative adversarial network (GAN) to learn student network.  Belagiannis \emph{et al.}~\cite{belagiannis2018adversarial} even studied the distillation of dense features using GANs. Since, teacher can transfer limited amount of knowledge to student, so Mirzadeh \emph{et al.}~\cite{mirzadeh2019improved} propose multi-step knowledge distillation, which employs intermediate-sized networks. Peng \emph{et al.}~\cite{peng2019correlation} propose a framework named correlation congruence for knowledge distillation (CCKD), which transfers the sample level knowledge, yet in addition, it also transfers the correlation between samples. 

  Few studies that utilize the dense features along with soft-logits claim that the dense features help to generalize the student model~\cite{zhu2021complementary,liu2020adaptive,li2020hierarchical}. Heo \emph{et al.}~\cite{heo2019comprehensive} set out a novel feature distillation technique in which the distillation loss is intended to make an alliance among different aspects: teacher transform, student transform, distillation feature position and distance function. An online strategy has also been proposed that eliminates the need for a two-phase strategy and performs joint training of a teacher network as well as a single multi-branch student network~\cite{NIPS2018_7980}. All these methods, including multi-branch strategy~\cite{NIPS2018_7980}  train only a single student on the final logits; we instead train multiple students, each on a chunk of dense representation. \\

\noindent
\textbf{Transferring knowledge to multiple student}: 
The only know method that uses multiple students is proposed by You \emph{et al.}~\cite{you2018learning}. They learned multiple binary classifiers (gated Support Vector Machines) as students from a single teacher which is a multi-class classifier. There are three problems with this approach. Firstly, as the number of classes in the dataset increases, the number of students required would also increase i.e., $1000$ students would be required for $1000$ class classification problem; secondly, it is applicable only for the classification tasks; thirdly, even after the students have been trained, the output from the teacher network is needed at inference time. To the best of our knowledge no further work has been done in the Single Teacher Multi-Student domain; our proposed approach is the first CNN-based Single Teacher Multi-Student network, which, once trained, becomes teacher independent and has a wide variety of applications including,  classification. \vspace{-0.2cm} 
\section{Methodology}
\label{section:method}
\vspace{-0.2cm}
A large state-of-the-art network well trained for a certain problem is considered as a teacher, comparatively, a smaller network is deemed as a student. 
Unlike~\cite{hinton2015distilling,gao2021dag,gou2021knowledge} that employ a single student to extract knowledge from the teacher's soft logits; our proposed methodology transfers knowledge from dense representation and takes advantage of multiple students.
\subsection{Extracting dense representation from the teacher}
Neural networks typically produce a dense feature representation $\mathbf{d}$ which, in case of classification, is fed into class-specific neurons called logits, $z_i$ (the inputs to the final softmax). The “softmax” output layer then converts the logit, $z_i$ computed for each class into a probability, $\hat{y}$, defined as:
\begin{equation}
\hat{y} = \frac{exp(z_i/T)}{\sum_j^c{exp (z_j/T)}},
\end{equation}
where $c$ is the total number of classes in the dataset and $T$ is the temperature ($T > 1$ results in a softer probability distribution over classes~\cite{hinton2015distilling}). Usually, the teacher-student network minimizes the cross-entropy between soft targets $\hat{y}$ of the large teacher network and soft targets of the small student network. Since these soft targets $\hat{y}$  being the probabilities produced by the softmax on the logits $z_i$ contain the knowledge only about categorizing inputs into respective classes, learning these soft targets limits the student network to solving a specific problem, making it problem-centric. The general information about the dataset learned by a network is stored in the dense feature representation $\mathbf{d}$  that helps the student to better mimic the teacher and stabilizes the training~\cite{belagiannis2018adversarial}. It can be observed in a typical transfer learning scenario, where the logit $z_i$ is removed, and only the knowledge in the dense feature layer $\mathbf{d}$ is used to learn a new task by transfer of knowledge from a related task. For example, in the case of VGG-16, the output layer of 1000 class-specific logits is removed and the output of FC2-4096 is used for feature extraction. Similarly, in case of ResNet34~\cite{he2016deep} and GoogLeNet~\cite{szegedy2016rethinking}, FC1-512 and Flattened-1024 are used for feature extraction respectively. 

Thus, we redefine the goal of knowledge transfer to that of training a small student model to learn the space spanned by a large pre-trained teacher network. This is achieved by minimizing the reconstruction error between the dense feature vector of the teacher network and the one produced by the student network as shown below:
\begin{equation}
      L(\mathbf{d},\hat{\mathbf{d}}) = \frac{1}{m} \sum_{i=1}^{m}(\mathbf{d}-\hat{\mathbf{d}})^2,
      \label{eq:mse1}
\end{equation}
where $m$ is the total number of training samples, $\mathbf{d}$  and $\hat{\mathbf{d}}$ are the dense feature representation of teacher and student networks, respectively. The dense representation $\mathbf{d}$ is obtained from a teacher network by extracting the output of the layer before the logits layer. Once student has learned to reconstruct dense representation, the output layer (e.g., class-specific logit and softmax in case of a classification problem) can be introduced to obtain the desired output as in transfer learning. This output layer could be the teacher's output layer with pre-trained weights. The same strategy can be extended to multiple student networks (\emph{teacher-class}) where the dense feature representation $\mathbf{d}$ can then be divided into multiple chunks (sub-spaces); and each sub-space can be learned by an independent student model.%; as discussed in the following sections.
% ======================================
%
%
\begin{table}
%\begin{adjustbox}{width=\textwidth}
\caption{Knowledge distillation error $(L_k)$ of each student while learning knowledge sub-spaces using only MSE loss and MSE + GAN loss in $S^4$ configuration.\\}
\centering
\resizebox{0.85\textwidth}{!}{%
\begin{tabular}{l|cccc|cccc}
 \hline
 & \multicolumn{4}{c|}{MSE} & \multicolumn{4}{c}{MSE + GAN}\\
 \cline{2-9}
 Dataset & \textbf{ $S^4_1$} & \textbf{$S^4_2$} & \textbf{$S^4_3$} & \textbf{$S_4^4$}& \textbf{ $S^4_1$} & \textbf{$S^4_2$} & \textbf{$S^4_3$} & \textbf{$S_4^4$}\\
 \hline
 MNIST &0.343  &0.393 &0.395 &0.368 &0.279 & 0.312 & 0.314 & 0.347 \\
 F-MNIST &0.157 &0.176 &0.173 &0.174&0.546 &0.704&0.58&0.546\\
 IMDB Movie Reviews &0.004  &0.003 &0.002 &0.002&-  &- &- &-\\
 \hline
\end{tabular}}
%\end{adjustbox}
\label{table:mse}
\end{table}
%
%\subsection{Mapping dense representation on vector spaces and sub-spaces}
%The dense representation $\mathbf{d}$ extracted from the teacher network is a vector space, and we can decompose this vector space into sub-spaces. Then, each sub-spaces are learned by a student network. After training the students, the learned sub-spaces are merged to produce original vector space. %Let $\mathbf{d}$ be the vector space we need to decompose.
%
\subsection{Learning dense representation using $n$ students}
Teacher-student methods~\cite{hinton2015distilling,mirzadeh2019improved,watanabe2017student} attempt to distill knowledge using one student, which becomes cumbersome for a simple network. Multiple students can also be utilized to mimic the teacher's knowledge. A previous such attempt resulted in an ensemble of binary classifiers~\cite{you2018learning}. In case of $1000$ class classification such as ImageNet~\cite{krizhevsky2012imagenet} this will require $1000$ student networks making the solution impractical for larger datasets.

Instead, in our case, the dense features (vector space) $\mathbf{d}$ is divided into certain number of sub-spaces, each containing partial knowledge. The vector space could be split into $n$ mutually exclusive sub-spaces or by using standard vector factorization methods such as singular value decomposition. The latter becomes impractical when the dataset has large numbers of examples. So, we simply split the $\mathbf{d}$ vector space into $n$ non-overlapping sub-spaces as $\mathbf{d} = [d_1,\textbf{0},...] + [\textbf{0},d_2,\textbf{0},...] + ... + [\textbf{0},...d_n]$ where each $d_k$ would be learned independently by the $k^{th}$ student. 
Let's assume that we have a set of $n$ students such that $S^n = \{S^n_k \mid k \in \mathbb{Z} \land 1 \leq k \leq n	\}$, where $S^n_k$ is $k^{th}$ student in the set. Mathematically, transferring the knowledge from teacher to $n$ students can be defined as:
\begin{equation}
\label{eq:student}
    \hat{d_k} = S_k^n (\mathbf{X}, \theta_s^k),  \text{where k} = 1, ..., n \\
\end{equation}
where, $\hat{d_k}$ is knowledge distilled by $k^{th}$ student by simply distillation loss or adversarial loss (in case the problem is addressed through adversarial learning). The loss for both the training methods is given in Table~\ref{table:mse} for three datasets where different loss for identical students indicates that learning each sub-space converges at different point. 
\begin{figure}[t!]
    \centering
    \includegraphics[width=.75\columnwidth]{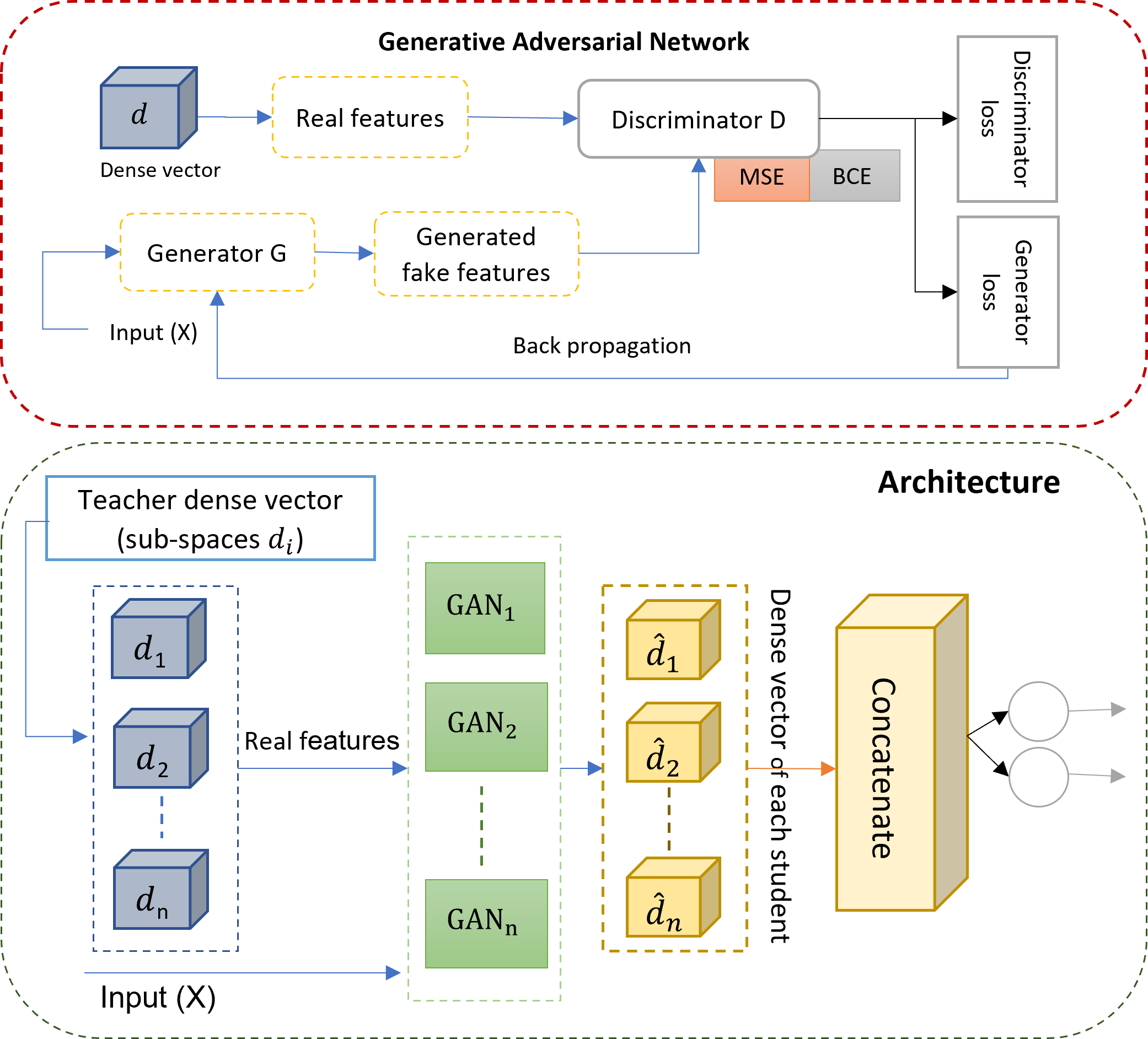}
    \caption{GANs' architecture diagram showing dense vector $\mathbf{d}$ produced by pre-trained Teacher (T) from input image $\mathbf{X}$ and fake samples generated by Generator {G}. The Discriminator {D} then uses binary cross entropy (BCE) to discriminate between real and fake feature vectors as well as mean squared error (MSE) to ensure reconstruction of dense representations. In multi-student configurations the the dense vector $\mathbf{d}$ is first divided into sub-spaces and then each student generates its respective sub-space $d_i$.}
    \label{fig:gans_diagram}
    \vspace{-.2cm}
\end{figure}

\subsection{Combining learned sub-spaces}
\vspace{-0.2cm}
% Combining $n$ Expert Students to form a single combined network
After all the $n$ students are trained independently, the learned sub-spaces $(\hat{d_k})$ are merged together to estimate the knowledge $(\hat{\mathbf{d}})$ learned by all $n$ students and is defined as:
\begin{equation}
\label{eq:student_z}
      \hat{\mathbf{d}} = S^n_1(\mathbf{X})+S^n_2(\mathbf{X})+ ... + S^n_n(\mathbf{X})
\end{equation}
where $\mathbf{X}$ is input data (e.g. images). Since the students have now collectively learned the space spanned by teacher, so they should ideally give the same results as the teacher network (if solving the same problem the teacher was solving) when fed to the teacher's output layer. Thus, in case of classification, the softmax layer can generate the probability vector as:
\begin{equation}
\label{eq:student_f}
      \hat{y} = g([S^n_1(\mathbf{X})+S^n_2(\mathbf{X}) + ... + S^n_n(\mathbf{X})], \theta_g)
\end{equation}
where the function $g$ represents the output layers (softmax in case of classification) applied on concatenation of output from all pre-trained students, $\hat{y}$ is predicted label, and $\theta_g$ are its weights which can also be pre-trained weights acquired from the teacher network. The knowledge learned by teacher i.e., $\mathbf{d}$ and $n$ students $\hat{\mathbf{d}}$ might have minor errors (see Table~\ref{table:mse}). To compensate this error and enhance the overall accuracy of the students, this output layer could be fine-tuned while keeping the students non-trainable. Thus, in case of classification, only last output layer can be optimized using cross-entropy loss function as:
\begin{equation}
      L_{class}(y,\hat{y}) =\sum_{i=0}^{N} y_i log(\hat{y}_i).
      \label{eq:CCE}
\end{equation}
%
% Table 5
\begin{table}[t]
%\begin{adjustbox}{width=8.0cm}
%
\caption{The impact of fine-tuning output layers $(g)$ on FF in configuration i.e $S^4$.\newline}
\centering
\resizebox{0.7\textwidth}{!}{%
\begin{tabular}{lcc}
 \hline
 Dataset &Without Fine-tuning & With Fine-tuning \\
 \hline
 MNIST   &97.66\%  &97.81\% \\
 %\hline
 {Fashion MNIST} &85.70\%  &86.54\%  \\
 %\hline
IMDB Movie Reviews  &88.29\%  &88.34\% \\
\hline
\end{tabular}
}
%\end{adjustbox}
\label{table:finetune}
\end{table}
\vspace{-0.3cm}
\subsection{Mapping as reconstruction problem}
\vspace{-.1cm}
Inspired by the success of generative adversarial networks (GANs) for solving sub-optimal problems of learning distributions~\cite{chang2020tinygan,wang2018adversarial, gou2021knowledge,gao2021dag}, we pose the dense representation learning as generative adversarial problem. Features from pre-trained Teacher (T) are considered as the real data distribution and they are mimic by student generative networks ({G}) as fake distribution where discriminator ({D}) distinguishes between the real and fake features.  Unlike GANs based distillation~\cite{wang2018adversarial,belagiannis2018adversarial},  several students are schooled in the adversarial fashion (see Fig.~\ref{fig:gans_diagram}). The choice of cross-entropy loss in discriminator provides information only about the real or fake sample based upon where the point lies relative to the decision boundary. However, the distance from the decision boundary would further penalize the generation of dense representation. Therefore, along with adversarial loss, we try to minimize the distance between real and fake samples through MSE loss.% as shown in Fig.~\ref{fig:gans_diagram}.
\vspace{-.2cm}
\section{Evaluation and Results}
\vspace{-.2cm}
We compare the proposed multi-student \emph{(teacher-class)} approach with well-known \emph{teacher-student} architectures~\cite{hinton2015distilling,wang2018adversarial,tian2019contrastive,guo2020online,NEURIPS2018_019d385e}. For a fair comparison, we keep the total number of parameters in all students combined equivalent to or less than a student in the \emph{teacher-student} approach for one set of experiments. Therefore, as $n$ increases, students become smaller and simpler. %For details of network architectures, please refer to the supplementary material. 
\subsection{Analysis of student population}
We analyzed the effect of increasing the number of students by designing different student architecture using both feed-forward (FF) and GAN-based strategy. In order to clearly study the effect of knowledge distillation via dense vectors, we kept our comparison with vanilla knowledge distillation~\cite{hinton2015distilling} which uses logits-based cost function. As shown in Table~\ref{table:accuracy1GAN}, for all four configurations on all datasets, the proposed method achieves accuracy comparable to their teacher network and higher than knowledge distillation (KD)~\cite{hinton2015distilling}. Overall, GAN-based distillation attains better accuracy than FF due to adversarial and distillation MSE losses. The single student created using vanilla KD approach~\cite{hinton2015distilling} has $6-19\%$ less accuracy than the teacher, whereas through our approach, using FF, $S^1$ performs better and using GANs, all $S^k$ configurations have equal or better accuracy on MNIST dataset. For Fashion MNIST (F-MNIST) poor performing student configuration is within $1\%$ of teacher's accuracy. For CIFAR-10, $S^{KD}$ has almost $10\%$ less accuracy than the teachers, whereas our student configuration attain within $2-7\%$ of the teacher's accuracy. 
\begin{table*}[t]
  \centering
  \caption{Comparison of knowledge distillation $(S^{KD})$~\cite{hinton2015distilling}, and KDGAN~\cite{NEURIPS2018_019d385e} with the proposed feed-forward (FF) and GAN based methods in terms of test accuracy reported on two different tasks and four datasets. $T$ is the teacher network, $S^n$ is $n$ student configuration of our proposed approach.\\}
  \resizebox{\textwidth}{!}{
  \def\arraystretch{1.5}
  \begin{tabular}{lc|cc|cccccccc}
    \hline
    \multirow{2}{1cm}{{Datasets}} &
    \multirow{2}{1cm}{{T}} &
    {{$S^{KD}$}} &
    {{KDGAN}} &

    \multicolumn{2}{c}{{$S^1$}} & 
    \multicolumn{2}{c}{{$S^2$}} &
    \multicolumn{2}{c}{{$S^4$}} &
    \multicolumn{2}{c}{{$S^8$}} \\
    % \hline
    \cline{5-12}
    &&\cite{hinton2015distilling}&{\cite{NEURIPS2018_019d385e}}& 
    {FF} & {GAN} & {FF} & {GAN} & 
    {FF} & {GAN} & {FF} & {GAN} \\
    % \hhline{~--}
    \hline
    MNIST & 98.11\% &93.34\% & 99.25\% &98.65\% &99.16\% & 96.79\% & \bf{99.30}\% &97.81\% &99.21\%&93.97\%&98.10\% \\ 

    F-MNIST & 91.98\% &82.87\% &-&89.97\% &89.35\% & 89.43\% & \bf{90.74}\% &86.54\% &89.49\%&82.33\%&90.67\%  \\ 

    CIFAR-10 & 89.85\% &79.99\% &86.50 &82.17\% &81.74\% & 82.32\% & 84.70\% &81.57\% &86.04\%&81.91\%&\bf{86.96}\% \\ 
    IMDB & 86.01\% &67.78\% &-&\bf{84.58}\% &84.48\% & 83.01\% & 83.28\% &83.29\% &83.28\%&83.58\%&83.67\% \\ 
    %CamVid & 44.7\% & 32.72\% & & &- & 33.4\% & & - & & - & \\
    \hline
  \end{tabular}}
 
  \label{table:accuracy1GAN}
\end{table*}
\subsection{Identical vs. non-identical students}
The convergence of each student at different loss values indicates that all sub-spaces are distinct; therefore, spaces hard to learn may require a better student network. Once all identical students converged, we improved low-performing students. This could be done either by increasing model depth (layers) or width (filters in layers); we followed the prior strategy. As shown in Table~\ref{table:mse}, for MNIST $S^4_2$ and $S^4_3$, for F-MNIST  $S^4_2$, $S^4_3$, $S^4_4$, and for IMDB $S^4_1$ and $S^4_2$ have relatively higher error. { {Therefore, by improving these, the student $S^4_3$ showed better error on MNIST and F-MNIST datasets whereas $S^4_1$ showed better error on IMDB dataset.}} Consequently, the error for learning the space by $n$ students combined also improved for all datasets. Overall, enhancing the weaker students ameliorate the performance at the cost of some additional computation due to increased parameters.
\begin{table*}[t]
%\begin{adjustbox}{width=8.0cm}
\caption{The impact of improving poor performing student in $(S^4)$ configuration in terms of  knowledge transfer error (MSE). Dataset with $^\dagger$ symbolizes the improved students.\\}
\centering
\resizebox{1\textwidth}{!}
{%
\begin{tabular}{l|cc|cc|cc|cc|cc}
 \hline
 \multirow{2}{*}{Dataset} &  \multicolumn{2}{c|}{\textbf{ $S^4_1$}} &  \multicolumn{2}{c|}{\textbf{$S^4_2$}} &  \multicolumn{2}{c|}{\textbf{$S^4_3$}} &  \multicolumn{2}{c|}{\textbf{$S^4_4$}} & \multicolumn{2}{c}{$\sum_{i=0}^{4}${$S^4_i$}} \\
 \cline{2-11}
 & MSE & \#Para & MSE & \#Para & MSE & \#Para & MSE & \#Para & MSE & \#Para  \\ 
 \hline
  MNIST &0.3432 &22.17k &0.3930 &22.17k &0.3948 &22.17k &0.3682 &22.17k & 1.4994 &88.68k \\
  MNIST$^\dagger$ &0.3432 &22.17k &0.2984 &85.74k &0.2976 &85.74k &0.3682 &22.17k & 1.3074 &215.82k \\
 \hline
 F-MNIST &0.1571 &22.17k  &0.1759 &22.17k &0.1728 &22.17k &0.1743 &22.17k & 0.6801 &88.68k \\
 F-MNIST$^\dagger$ &0.1571 &22.17k &0.1293 &85.74k &0.1112 &85.74k &0.1131 &85.74k & 0.5107 &278.39k \\
 \hline
 IMDB Movie Reviews &0.0043  &165.45k &0.0030 &165.45k &0.0024 &165.45k &0.0019 &165.45k & 0.0116 &661.6k \\
 IMDB Movie Reviews$^\dagger$ &0.0017 &331.6k &0.0021 &331.6k &0.0024 &165.45k &0.0019 &165.45k & 0.0081 &994.1k \\
 \hline
\end{tabular}}  
%\end{adjustbox}
%
\label{table:nonidc}
\end{table*}
\subsection{Fine-tuning student networks}
As discussed in section~\ref{section:method}, once students have been trained, the knowledge learned by all students is combined and fed to an output-specific layer. If the task for students is the same as that of the teacher, then output-specific layers can be borrowed from the teacher network and initialized with pre-trained weights. These layers may or may not need fine-tuning, although it would improve the performance in some cases. To study the effect of using pre-trained layers $(g)$ with or without fine-tuning, experiments were performed on $4$ student configurations. From Table~\ref{table:finetune}, it can be observed that for MNIST and F-MNIST, there is an improvement of approximately $1\%$ in test scores. For the IMDB Movie Reviews dataset, it is even less than $1\%$. This indicates that the dense representation $(\hat{\mathbf{d}})$ produced by all students together was already similar to the teacher's dense representation $(\mathbf{d})$.%, and the students had converged during their independent training.
\begin{table}[h]
%\begin{adjustbox}{width=8.0cm}

\caption{Comparison of knowledge distillation $(S^{KD})$~\cite{hinton2015distilling} with the proposed method in terms of inference time (in seconds) on slave machines within a network cluster for several configurations of the proposed approach.\\}
\centering
\resizebox{0.75\textwidth}{!}{%
\begin{tabular}{lcccccc}
 \hline
 Dataset & T & \textbf{$S^{KD}$} & \textbf{$S^2_1$} & \textbf{$S^4_1$} & \textbf{$S^6_1$} & \textbf{$S^8_1$}\\
 \hline
 F-MNIST &0.171s  &0.135s &0.096s &0.083s &0.082s &0.075s \\
IMDB Movie Reviews &0.162s &0.125s &0.056s &0.056s &0.053s &0.051s\\
 \hline
\end{tabular}}
%\end{adjustbox}
%
\label{table:cluster}
\end{table}

\begin{table*}[tbh]
\caption{The comparison of network size (parameters) and FLOPS of Teacher, one student(~\cite{hinton2015distilling}), and several configurations of the proposed method. The graph on the right shows compression of the student with reference to the teacher as we increase the number of students. The values of model size are normalized by the teacher's size.\\
}
\centering
\begin{adjustbox}{width=0.95\textwidth}
\resizebox{\columnwidth}{!}
{%
\begin{tabular}{c|cc|ccc}
\cline{1-5}
\multirow{2}{*}{Config.} & \multicolumn{2}{c|}{MNIST \& F-MNIST} & \multicolumn{2}{c}{IMDB Movie Reviews}  & \multirow{11}{*}{\includegraphics[width=0.65\columnwidth]{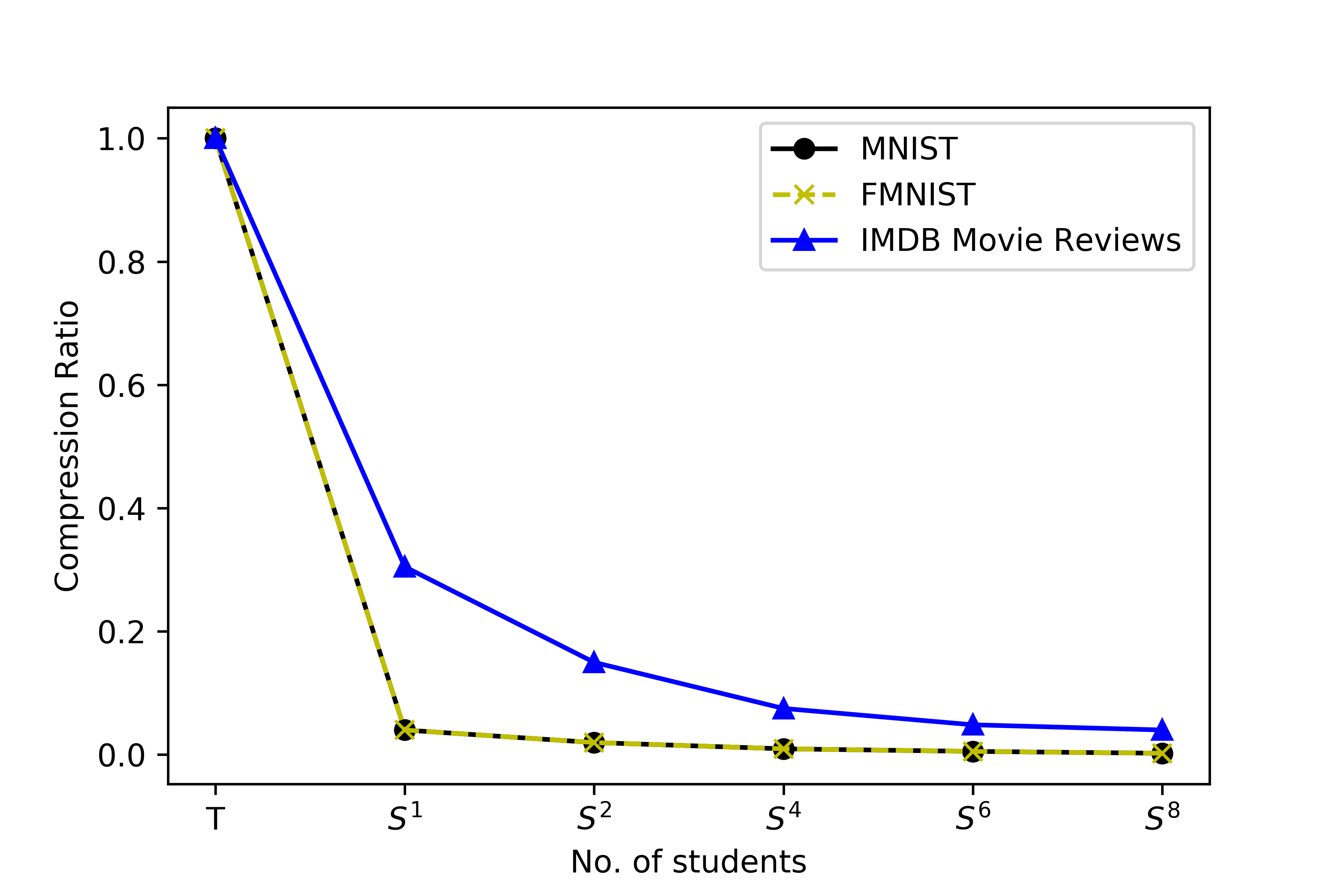}}\\
\cline{2-5}
& \#Para & FLOPs & \#Para & FLOPs  \\
\cline{1-5}
Teacher & 2.38M & 26.46M & 2.21M & 7.20M \\
\cline{1-5}
$S^{KD}$ & 94.65k &11.67M & 673.80k & 2.57M   \\
\cline{1-5}
$S^2$ & 95.32k &12.53M & 662.03k & 2.23M   \\
$S^2_k$ & 46.36k &6.26M & 330.88k & 1.11M   \\
\cline{1-5}
$S^4$ & 91.24k &12.54M & 662.05k & 2.25M   \\
$S^4_k$ & 22.17k &3.14M & 165.45k & 562.11k   \\
\cline{1-5}
$S^6$ & 67.84k &9.46M & 642.48k & 1.49M   \\
$S^6_k$ & 10.9k &1.57M & 106.8k & 249.2k   \\
\cline{1-5}
$S^8$ & 41.37k &1.19M & 705.3k & 913.02k   \\
$S^8_k$ & 5.00k &149.12k & 88.13k & 114.06k   \\
\cline{1-5}
\end{tabular}}
\end{adjustbox}
\label{table:model_size}
\end{table*}
%
%\vspace{-0.4cm}
\subsection{Computational cost}
%\vspace{-0.2cm}
%To prove the efficacy of parallelism and small multiple students, 
While designing students, the total number of parameters in the multi-student framework was kept equivalent to one student setup~\cite{hinton2015distilling}, as shown in Table~\ref{table:model_size}. Such as, for MNIST and F-MNIST, the teacher has $11.67M$ parameters,  and one student~\cite{hinton2015distilling} has $94.65$k. While, each student in the $2$, $4$, $6$ student configuration of our approach has $46.36$k, $22.17$k, $10.9$k parameters, respectively. Thus, when 8 student configuration $S^8_1$ is used, the individual student becomes as small as just $5000$ parameters, which makes training a model much easier. Similarly, for the IMDB Movie Review dataset, one student of $673.80$k parameters was halved to $330.88$k parameters to design two students and quartered to $165.45$k parameters to create four students. Effectively, the decline in model size also reduced FLOPS per student. The graph adjacent to Table~\ref{table:model_size} demonstrates the normalized model size of a single student concerning the teacher network. Here, the teacher and the student model for MNIST and F-MNIST datasets were the same. It can be observed that as we increase the number of students, the individual student becomes smaller.    
%
% ================================================
\vspace{-0.4cm}
\subsection{Inference time on distributed cluster}
We tested $n$ students on a cluster of virtual machines (VM's) for two datasets where student-$n$ would run on slave-$n$. The master computer would ask the slave computers to infer a sample of data and send back the results to it. The weights and sample data each student requires for inference were all placed in a shared folder such that they were accessible to all systems within the cluster. The detail of the cluster is discussed in the supplementary material. The average round trip time for our virtual cluster was 0.0365 milliseconds. A teacher and a student of the knowledge distillation approach were executed on a single slave system to compute their inference times. As evidenced in Table~\ref{table:cluster}, our proposed methodology outperforms the teacher and the single student of the KD~\cite{hinton2015distilling} approach in terms of inference time with the added benefit that students can execute in parallel in a cluster.
\vspace{-0.3cm}
\subsection{Results on ImageNet}
%\vspace{-0.2cm}
{To prove the efficacy of the proposed method, we compare it with seven up-to-date approaches~\cite{guo2020online,li2020local,du2020agree,tian2019contrastive, zhu2021complementary, belagiannis2018adversarial, li2020hierarchical} on the ImageNet dataset}. We employed ResNet-50 pre-trained as a Teacher, ResNet-18 as one student (FF-$S^1$), and ResNet-9 (removing the recurring residual block in ResNet-18) in two student setup (FF-$S^2$). Table~\ref{table:imagenet} depicts that using FF training with only mse loss based distillation. Our method is better/comparable than four of the methods (AEKD~\cite{guo2020online}), AVER~\cite{guo2020online}, ANC~\cite{belagiannis2018adversarial}, HKSANC~\cite{li2020hierarchical}) while four methods outperform  our approach (Online KD via CL~\cite{guo2020online}, LC KD~\cite{li2020local}, CRD~\cite{tian2019contrastive}, CRCD~\cite{zhu2021complementary}). Improved performance of Online KD via CL~\cite{guo2020online} is due to the use of information fusion in an online manner. Similarly, to supplement distillation loss, LC KD~\cite{li2020local} uses local correlation where they apply knowledge distillation on the hidden layers in addition to soft labels. Likewise, CRD~\cite{tian2019contrastive} and CRCD~\cite{zhu2021complementary} uses additional data correlation information for knowledge transfer. It should be noted that our approach offers a new strategy for distillation and many advancements proposed over vanilla KD could be employed to further improve each of our students.   

%, our method achieves comparable results despite floating point errors. Our teacher's low accuracy and training on Google TPU using $bf16$ instead of $float32$ may have resulted in some reduction in performance. 

%
\begin{table*}[h]
\centering
%\begin{adjustbox}{\columnwidth}
\caption{Comparison with state-of-the-art on ImageNet dataset using ResNet-50 as teacher and ResNet-18 as student.  \newline}
\resizebox{0.65\textwidth}{!}{%
%\vspace{11}
\begin{tabular}{l|cc|cc}
\hline
\multirow{2}{*}{Method} & \multicolumn{2}{c|}{Baseline/Teacher's Acc.} & \multicolumn{2}{c}{Student's Acc.} \\ 
 & Top-1$\uparrow$ & Top-5$\uparrow$ & Top-1$\uparrow$ & Top-5$\uparrow$ \\ \hline
Online KD via CL~\cite{guo2020online} & 77.8\% & -  & 73.1\% & - \\ %\hline
LC KD~\cite{li2020local} & 73.27\% & 91.27\% & 71.54\% & 90.30\% \\ %\hline
AE KD~\cite{du2020agree} & 75.67\% & 92.50\% & 67.81\% & 88.21\% \\ %\hline
AVER~\cite{du2020agree} & 75.67\% & 92.50\% & 67.81\% &  88.21\% \\ %\hline
CRD~\cite{tian2019contrastive} & 73.31\% & 91.42\% & 71.38\% & 90.49\% \\%\hline
CRCD~\cite{zhu2021complementary} & 73.31\%
 & 91.42\% &  71.96\%  & 90.94\%\\%\hline
ANC~\cite{belagiannis2018adversarial} & 72.37\% & 94.1\% & 67.11\% & 88.28\% \\%\hline 
HKSANC~\cite{li2020hierarchical} & - & - &  68.66\%  & 89.15\%\\%\hline 
FF-$S^1$ & 75.24\% & 92.19\% & 68.27\% & 88.81\% \\ %\hline
FF-$S^2$ & 75.24\% & 92.19\% & 66.32\% & 87.01\% \\ \hline
\end{tabular}
}
\label{table:imagenet}
%\end{adjustbox}
\end{table*}
%
%%%%%%%%%%%%
\subsection{Ease of training}
We divide a large and complex student model into multiple smaller and simpler students and learn them through distillation. Larger models require more data and compute for training, whereas simpler models could be trained in relatively fewer epochs~\cite{tan2021efficientnetv2}. From Table~\ref{table:EaseOfTraining}, it is clear that as we increase the number of students, each student model becomes smaller, the convergence is achieved faster and in fewer epochs. Such as using feed-forward $(FF)$ training, the total training time reduces from $80$ seconds to $30$ seconds in $S^2$ configuration and $15$ seconds in $S^4$ configuration on the MNIST dataset. Similarly, for Fashion MNIST, there is a decline in total training time while increasing the number of students. In the case of GANs, although there is a reduction in total training time, yet the change is relatively small compared to FF. Because the discriminator model in GANs reduces the effect of model simplicity during training time. Nevertheless, the GANs-based trained students perform better and can compete with FF during inference time. In a nutshell, the distillation of knowledge to multiple students makes each student simple, small, benefits from parallelism, and minimizes the overall training time.         

\begin{table*}[!htb]
\centering
%\begin{adjustbox}{\columnwidth}
\caption{{Table showing ease of training and benefit of parallelism via multi-student distillation. The training converges in same or fewer epochs, and per epoch computation time as well as total training time reduces.}}

\resizebox{0.8\textwidth}{!}
{
\begin{tabular}{l|rrrrrrr}
\hline%\toprule
&{Dataset} &{TCN} &{Epochs} &{Time/epoch} &{FLOPS} &{Total time} \\
& & & &(sec) & & (sec) \\\hline
\multirow{6}{*}{FF} &\multirow{3}{*}{MNIST} &$S^1$ &20 &4 &2.9M &80 \\
& &$S^2$ &10 &3 &2.9M &30 \\
& &$S^4$ &10 &1.5 &1.5M &15 \\
\cmidrule{2-7}
&\multirow{3}{*}{F-MNIST} &$S^1$ &10 &4 &2.9M &40 \\
& &$S^2$ &10 &3.5 &2.9M &35 \\
& &$S^4$ &10 &3 &1.5M &30 \\
\hline
\multirow{6}{*}{GANS} &\multirow{3}{*}{MNIST} &$S^1$ &40 &14 &2.9M &560 \\
& &$S^2$ & 35 &12.5 &2.9M &437.5 \\
& &$S^4$ & 35 &12.5 &1.5M &437.5 \\
\cmidrule{2-7} 
&\multirow{3}{*}{F-MNIST} &$S^1$ &55 &13 &2.9M &715 \\
& &$S^2$ &55 &13 &2.9M &715 \\
& &$S^4$ &50 &13 &1.5M &650 \\
\hline%\bottomrule
\end{tabular}}
 \label{table:EaseOfTraining}
\end{table*}
%%%%%%%%%%%%

\vspace{-0.6cm}
\section{Conclusions}
\vspace{-0.1cm}
To transfer knowledge from a teacher to a student, we proposed a new method called \emph{teacher-class} network that 
decomposes the teacher's learned knowledge into sub-spaces, and unlike single teacher single student (STSS) architecture, it employs multiple students to learn the sub-spaces of knowledge. Rather than distilling logits, our method transfers dense feature representation that makes it problem independent, hence it can be applied to different tasks. Since the approach allows to train all students independently, therefore, these student networks can be trained on $CPU$ or even on edge devices over the network. Upcoming GPUs that support model parallelism at inference time, such as Groq, Habana Goya, Cerebras Systems, etc are best suited for our architecture. Through extensive evaluation, it has been demonstrated that the proposed method reduces the computational complexity, improves the overall performance, and outperforms the STSS approach. 
%\footnotesize
\bibliography{egbib}

% ##############################

\end{document}